\documentclass[11pt]{article}

\usepackage[preprint]{acl}

\usepackage{times}
\usepackage{latexsym}
\usepackage{amsmath}
\usepackage{amssymb}
\usepackage{graphicx}
\usepackage{subcaption}
\usepackage{fontawesome5} 

\usepackage[T1]{fontenc}
\usepackage[utf8]{inputenc}

\usepackage{microtype}

\usepackage{inconsolata}

\title{Beyond Model Bias: Cross-Entropy Based Rankings in Literary and Academic Texts Remain Stable across Small Language Models}

\author{Ankit Bhattacharjee$^{\dagger}$, Indrasabarna Hari$^{\star}$, Krityapriya Bhaumik$^{\dagger}$, Arka Das$^{\dagger}$, Piyush Sarangi$^{\dagger}$\\
$^{\dagger}$Indian Institute of Technology Kharagpur, India\\
$^{\star}$Asutosh College, University of Calcutta, India\\
\small \textbf{Correspondence:} 
\{\href{mailto:ankit2005@kgpian.iitkgp.ac.in}{ankit2005},
\href{mailto:ktpb124578@kgpian.iitkgp.ac.in}{ktpb124578}, \href{mailto:arkadas472@kgpian.iitkgp.ac.in}{arkadas472}, \href{mailto:piyushsarangi2013@kgpian.iitkgp.ac.in}{piyushsaranngi2013}\}@kgpian.iitkgp.ac.in, \href{mailto:scholarlystudies.indra@gmail.com}{scholarlystudies.indra@gmail.com}
}

\begin{document}
\maketitle

\begin{center}
\faGithub\ \small \textbf{Data \& Code:}\\ \url{https://github.com/SrabonGitikar/Cross-Entropy}
\end{center}

\vspace{0.2 em}

\begin{abstract}
While language models provide powerful mechanisms for evaluating textual predictability, it remains unclear whether the relative extents of LM surprisal remain stable across models. This study presents empirical evidence of the invariance of ranking based on cross-entropy related measures (e.g. mean, variance of each data-point), across four distinct small scale transformer-based models, for a given dataset of text sources. We consider two stylistic environments: a corpus of contemporary poetry and a corpus of academic research papers. We demonstrate that mean and variance of cross-entropy maintain robust cross-model rank correlation ($\rho > 0.90$) across both regimes. To test whether this agreement merely reflects generic differences in textual complexity, we compare it against six parameter-free compression baselines. The extent of agreement among language models is found to be substantially higher than that among model-baselines, for most of the measures. Geometric analysis of the structurally constrained academic regime reveals that higher-order moments are sensitive to specific models, although strong agreement persists. The findings suggest that relative extents of textual unpredictability may be independent of the language model they are computed against.
\end{abstract}

\section{Introduction}

Quantitative stylometry and computational text analysis rely on statistical representations to capture variations in style, register, and structural complexity across human writing \cite{holmes1985analysis, evert2017understanding}. Language model cross-entropy (or token surprisal) has emerged as an information-theoretic lens for quantifying textual unpredictability, capturing how unexpected a text sequence is under a learned probability distribution \cite{hale2001probabilistic, levy2008expectation, sarfati2025prompt}. However, raw surprisal values are tied to an evaluation model's specific vocabulary size, tokenizer granularity, and pre-training corpus. This architectural dependency creates a fundamental ambiguity in computational linguistics: does the relative order of language model cross-entropy metrics remain stable across models, or are they artifacts of the specific model? 

To resolve this ambiguity, we investigate the variance of ranks of moments of cross-entropy across four architecturally distinct language models --- \texttt{GPT-2} \cite{radford2019language}, \texttt{Pythia} \cite{biderman2023pythia}, \texttt{TinyLlama} \cite{zhang2024tinyllama}, and \texttt{Qwen} \cite{qwen2023}. We contrast two structural bounds of language: an \textit{unconstrained regime} of contemporary poetry (28 poets) and a \textit{constrained regime} of academic prose (29 research papers). Combined, the two datasets comprise $\sim3,000$ total chunks of texts (the mechanism of chunking is given in appendix 
\ref{app:chunk}), which include $1,597$ poetries of the contemporary period and 29 recently published research papers across various disciplines,  for which cross-entropies are computed. Given the drastic differences in subword tokenization boundaries, vocabulary capacities, and pre-training corpora among these models, source-level (a source is a poet, or a paper, depending on which corpus is being referred to) surprisal rankings are theoretically vulnerable to variability. However, we empirically demonstrate that the relative statistical moments of textual unpredictability---mean, variance, skewness, and kurtosis---remain highly stable. These findings suggest that the extent of language model surprisal is not a model-specific artifact, but a model-independent feature of human text registers.

To summarize, main contributions of this study are as follows:
\begin{itemize}
    \item \textbf{Empirical Evidence of Rank-Invariance across Models:} We demonstrate that, for literary and academic texts, the ranks of statistical moments of language model surprisal (mean, variance, skewness, and kurtosis) are almost invariant across some distinct Transformer architectures, surviving massive disparities in vocabulary size and pre-training distributions of different language models. To demonstrate robustness, we conduct the study with multiple poets and multiple research papers across different disciplines. 
    \item \textbf{Contrastive Regime Analysis:} We evaluate this stability across two structural extremes --- an unconstrained creative regime (28 contemporary poets) and an academically constrained regime (29 research papers) --- proving that relative stylometric signal is robust across two different genres.
    \item \textbf{Potentially Invariant Stylometric Features:} Our findings therefore indicate that, though cross-entropy moments across models differ in their absolute values, they can be treated as objective, model-agnostic features for computational stylometry, through methods like relative ordering. This offers a robust, model-independent metric for downstream tasks such as genre classification and authorship attribution, though these remain outside the scope of the current study. The absolute values of cross-entropy do differ across different models, however the relative statistical measures, especially mean and variance of source-level cross-entropy remain largely invariant and maintain robust rank correlations across distinct language model architectures.
\end{itemize}

\section{Related Work}

\paragraph{Information-Theoretic Linguistics:} 
Language model surprisal, mathematically equivalent to token-level cross-entropy, is a foundational metric in psycholinguistics. Surprisal theory posits that the cognitive effort required to process a word is directly proportional to its unexpectedness in context \cite{hale2001probabilistic, levy2008expectation}. This framework underpins the Uniform Information Density (UID) hypothesis, which suggests that human communication naturally distributes surprisal evenly to maintain a constant information rate and avoid cognitive bottleneck spikes \cite{jaeger2010redundancy, zhan2021revisiting}. While constrained, formal text adheres closely to UID bounds, creative language—such as contemporary poetry—often deliberately violates these constraints, exhibiting high-variance surprisal gradients to achieve stylistic impact.

\paragraph{Computational Stylometry:} 
Stylometry quantifies these structural variations to classify text genre, register, and authorship. Traditional stylometric models rely on surface-level frequency distributions, such as syntactic dependencies, character $n$-grams, and Burrows' Delta \cite{holmes1985analysis, evert2017understanding}. As large language models (LLMs) have advanced, research has increasingly adopted deep contextual representations and next-token probabilities to map stylistic signatures \cite{sarfati2025prompt}. However, deploying language model surprisal as a feature for stylometry introduces a fundamental architectural bias. 

\paragraph{Model Bias \& Dependency:} 
The absolute magnitude of cross-entropy is strictly bound to a model's specific vocabulary capacity, tokenization granularity (e.g., BPE vs. SentencePiece), and pre-training distribution \cite{radford2019language, biderman2023pythia}. Because a single word may be split into drastically different subword sequences depending on the model, standard evaluation methodologies dictate that raw surprisal and perplexity cannot be directly compared across different tokenizer families without byte-level normalization \cite{meister2026whatlanguage}. 

\paragraph{Our Contribution and Novelty:} 
Our work breaks from the assumption that structural unpredictability is a model-bound artifact. While absolute cross-entropy values are undeniably tied to specific models, we establish that the \textit{ranks of statistical moments} (mean, variance, skewness, kurtosis) of source-level cross-entropies are stable across distinct language model architectures. By tracking these rank correlations across both an unconstrained poetic regime and a tightly constrained academic regime, we empirically demonstrate that the extent of language model surprisal for a text, survives model biases, particularly for small language models. Our core novelty is demonstrating that when properly normalized to character-level information density (BPC), the multi-model rank correlation ($\rho$) of cross-entropy moments remains stable across distinct model architectures.

\section{Methodology}
In this section, we formalize the empirical framework used to evaluate the model-based stability of textual unpredictability extents across distinct language model architectures. We rigorously define chunk-level cross-entropy as an empirical random variable, detail the extraction of its four statistical moments, and outline the non-parametric rank correlation metrics used to test structural stability across the constrained and unconstrained text regimes.

\subsection{Corpus Segmentation}

Let $S$ denote the complete set of textual sources, partitioned into two structurally distinct regimes: the poetic corpus $S_{\text{poet}}$ ($|S_{\text{poet}}| = 28$ poets) and the academic corpus $S_{\text{acad}}$ ($|S_{\text{acad}}| = 29$ papers). 

In each of the two cases, each source $i \in S$ is segmented into a set of contiguous text chunks. We define the chunk set for source $i$ as:
\begin{equation}
\mathcal{C}_i = \{c_{i,1}, c_{i,2}, \dots, c_{i,K_i}\},
\end{equation}
where $K_i$ represents the total number of chunks extracted from source $i$. This chunking is done so that for each source (i.e., for a poet in the first case and a paper in the second case), we obtain a frequency distribution of entropy for the source to ensure consistent evaluation across language models with finite context windows. The construction of a chunk for a given paper or a collection of poems written by a particular poet, is given in appendix~\ref{app:chunk}.

Because language models process discrete tokens rather than raw text, evaluation is strictly bound to a model's subword vocabulary. Let $\mathcal{M}$ represent the set of evaluated language models, where $\mathcal{M} = \{\text{\texttt{GPT-2}, \texttt{Pythia}, \texttt{TinyLlama}, \texttt{Qwen}}\}$. For any given model $M_k \in \mathcal{M}$, we define a distinct tokenization function $T^{(M_k)}$. Applying this function to a raw text chunk $c$ yields an ordered sequence of discrete tokens:
\begin{equation}
T^{(M_k)}(c) = (t_1, t_2, \dots, t_m),
\end{equation}
where the sequence length $m$ varies depending on the specific granularity and vocabulary size of model $M_k$.

\subsection{Distribution of Cross-Entropy}

For a given language model $M_k$, the probability of generating a specific token $t_j$ is conditioned on the preceding context $t_{<j}$. The model's internal representation of textual unpredictability is quantified by the token-level surprisal, defined as the negative log-likelihood of the observed token \cite{bengio2003neural}. 

We extend this to the chunk level by defining the cross-entropy $H^{(M_k)}(c)$ of a chunk $c$ in Bits Per Character (BPC). To ensure standardized evaluation across different subword vocabularies, we normalize the total negative log-likelihood of the chunk by its character length:
\begin{equation}
H^{(M_k)}(c) = -\frac{1}{n_c} \sum_{j=1}^{m} \log_2 P^{(M_k)}(t_j \mid t_{<j}),
\end{equation}
where $n_c$ is the total number of characters in the raw text chunk $c$. It is to be noted that the individual tokens are not normalized by their character counts, instead the normalization is performed after computing the summation.

Instead of reducing a source's entire corpus to a single scalar value, we evaluate the cross-entropy for every chunk in the set $\mathcal{C}_i$. This yields a discrete set of values $\{H^{(M_k)}(c_{i,1}), \dots, H^{(M_k)}(c_{i,K_i})\}$. We treat them as an empirical frequency distribution of textual unpredictability for source $i$. This frequency-based framing allows us to directly compute the structural volatility and statistical moments of the text for a given source (poet / paper).

\subsection{Moments and Their Rankings across Models}

To capture the complete stylistic fingerprint of a source, we extract the first four statistical moments from its empirical frequency distribution of cross-entropies: the mean ($\mu$), variance ($\sigma^2$), skewness ($\gamma$), and kurtosis ($\kappa$). 
We then test whether the relative ordering of sources, based on these four measures, remains stable across the set of language models $\mathcal{M}$. The agreement between two sets of ranked data is evaluated by computing Spearman's rank correlation coefficient ($\rho$) between pairs of models. The complete mathematical formulations are detailed in appendix \ref{app:moment}.

A correlation of $\rho \approx 1$ indicates that the language models, despite utilizing entirely different tokenization functions and pre-training distributions, extract the exact same relative stylistic ranking. This would indicate that cross-entropy based ranking may surpass the effect of model bias. 

Conversely, a result of $\rho \approx 0$ implies no correlation among the rankings of the models. That would indicate that textual unpredictability, when it is measured by language model cross entropy, is largely dependent on the specific model $M$ used to compute the cross entropy.  

\section{Experimental Setup}
Now, we formally describe the experiment performed to obtained the results that are described in the section \ref{sec:results}.

\subsection{Language Models}
To rigorously test the architectural independence of ranking based on moments of cross-entropy, we evaluate four distinct autoregressive language models. These models were selected specifically to maximize variance across model paradigms, vocabulary capacities, and pre-training distributions:
\begin{itemize}
    \item \texttt{GPT-2} \cite{radford2019language}
    \item \texttt{Pythia} \cite{biderman2023pythia}
    \item \texttt{TinyLlama} \cite{zhang2024tinyllama}
    \item \texttt{Qwen} \cite{qwen2023}
\end{itemize}

The models differ massively in both architecture and training corpora. The disparities are discussed thoroughly in appendix \ref{app:model}.

\subsection{Datasets}
We benchmark the stability of stylistic ranks across two extreme structural bounds:
\begin{itemize}
    \item \textbf{Poetic Regime ($S_{\text{poet}}$):} A corpus of contemporary poetry authored by 28 distinct poets. This dataset represents high stylistic volatility, where semantic and syntactic rules are frequently and intentionally subverted to achieve creative impact.
    \item \textbf{Academic Regime ($S_{\text{acad}}$):} A corpus of 29 published academic research papers. This dataset enforces strict structural homogeneity, shared technical terminology, and standardized formatting, severely compressing the baseline variance of textual unpredictability. The papers were deliberately chosen such that they contain minimum amount of mathematical equations and expressions. However we picked these papers from different disciplines. 
\end{itemize}

\subsection{Implementation}
All cross-entropy computations were performed using teacher forcing, without autoregressive generation. For each chunk $c_{i,j}$, the token sequence was processed by the corresponding language model $M_k$, and to ensure a standardized evaluation across different subword vocabularies, we normalize the total negative log-likelihood of the chunk by its character length, yielding the Bits Per Character (BPC) used in all subsequent moment calculations. Chunks exceeding the model's context window were processed using sliding windows, ensuring that each scoreable next token was evaluated once.

\section{Results \& Discussion}
\label{sec:results}
Now, we present the results and our findings across the two corpora that we experimented on.

\begin{figure*}[h!]
\centering
\begin{minipage}{0.75\textwidth} % Controls total footprint size
    \centering
    % Row 1: Mean and Variance
    \begin{minipage}{0.48\textwidth}
        \centering
        \includegraphics[width=\linewidth]{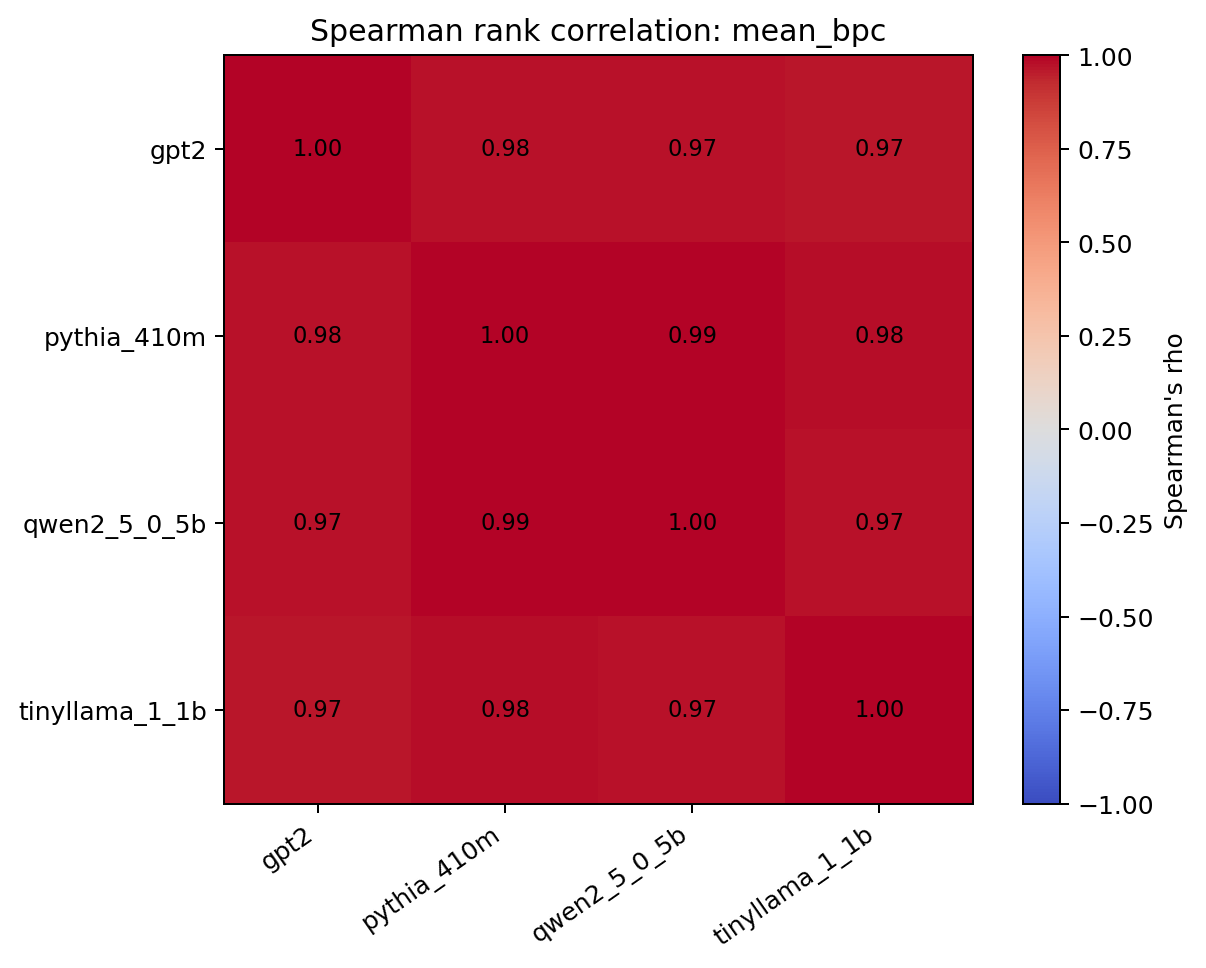}
        \subcaption{Mean ($\mu$)}
        \label{fig:heat_mean}
    \end{minipage}\hfill
    \begin{minipage}{0.48\textwidth}
        \centering
        \includegraphics[width=\linewidth]{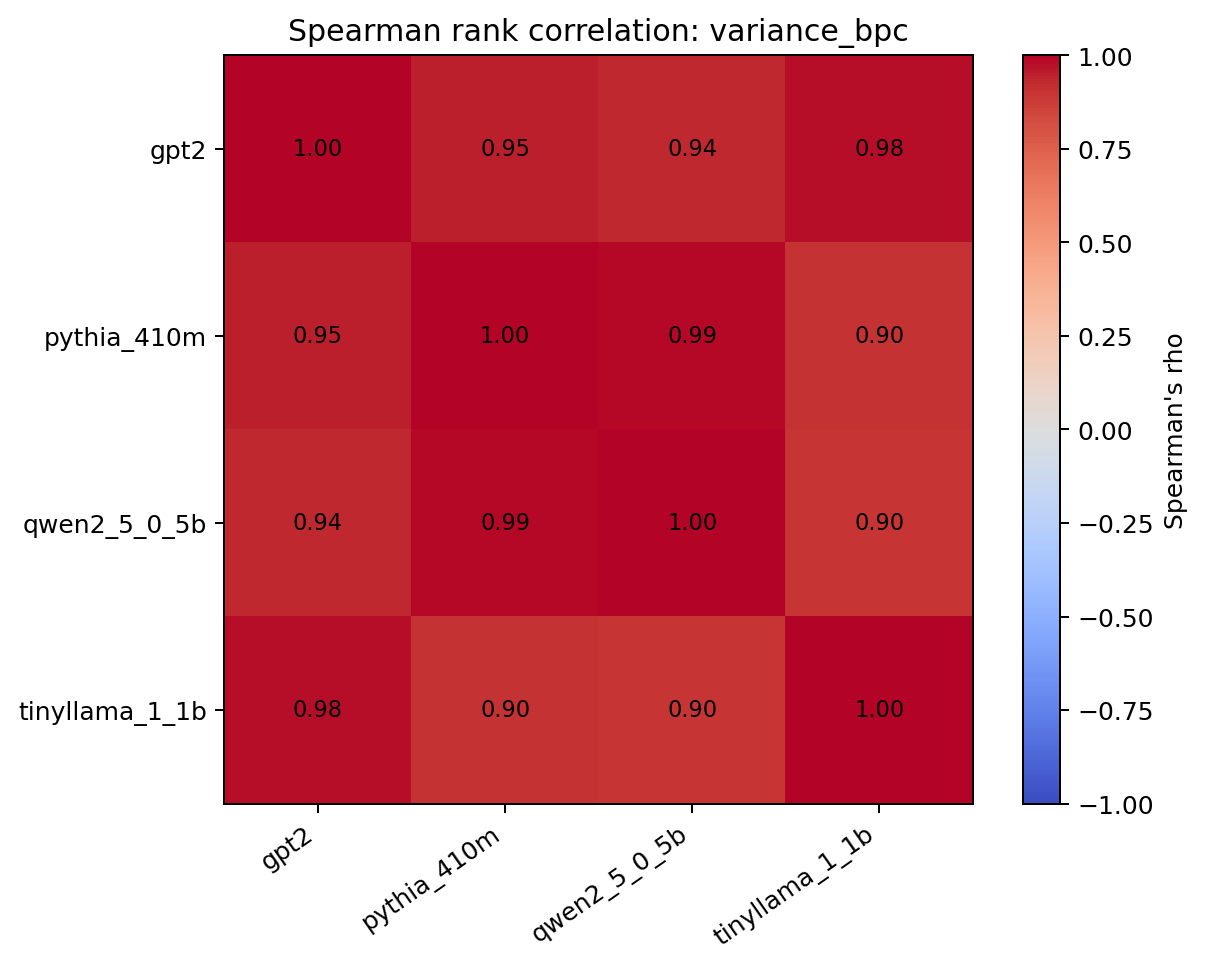}
        \subcaption{Variance ($\sigma^2$)}
        \label{fig:heat_var}
    \end{minipage}

    \vspace{0.4em} % Reduced vertical gap between rows

    % Row 2: Skewness and Excess Kurtosis
    \begin{minipage}{0.48\textwidth}
        \centering
        \includegraphics[width=\linewidth]{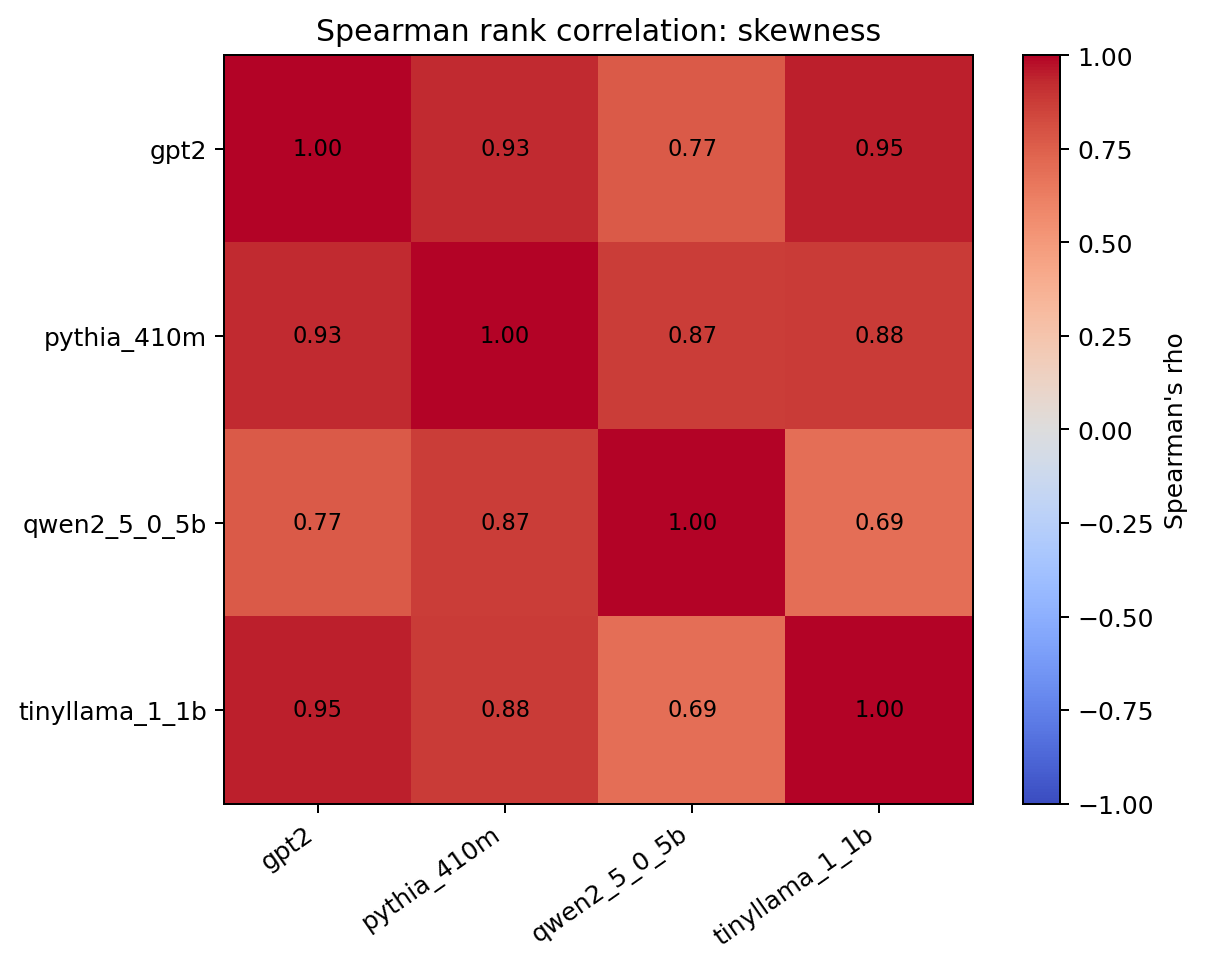}
        \subcaption{Skewness ($\gamma$)}
        \label{fig:heat_skew}
    \end{minipage}\hfill
    \begin{minipage}{0.48\textwidth}
        \centering
        \includegraphics[width=\linewidth]{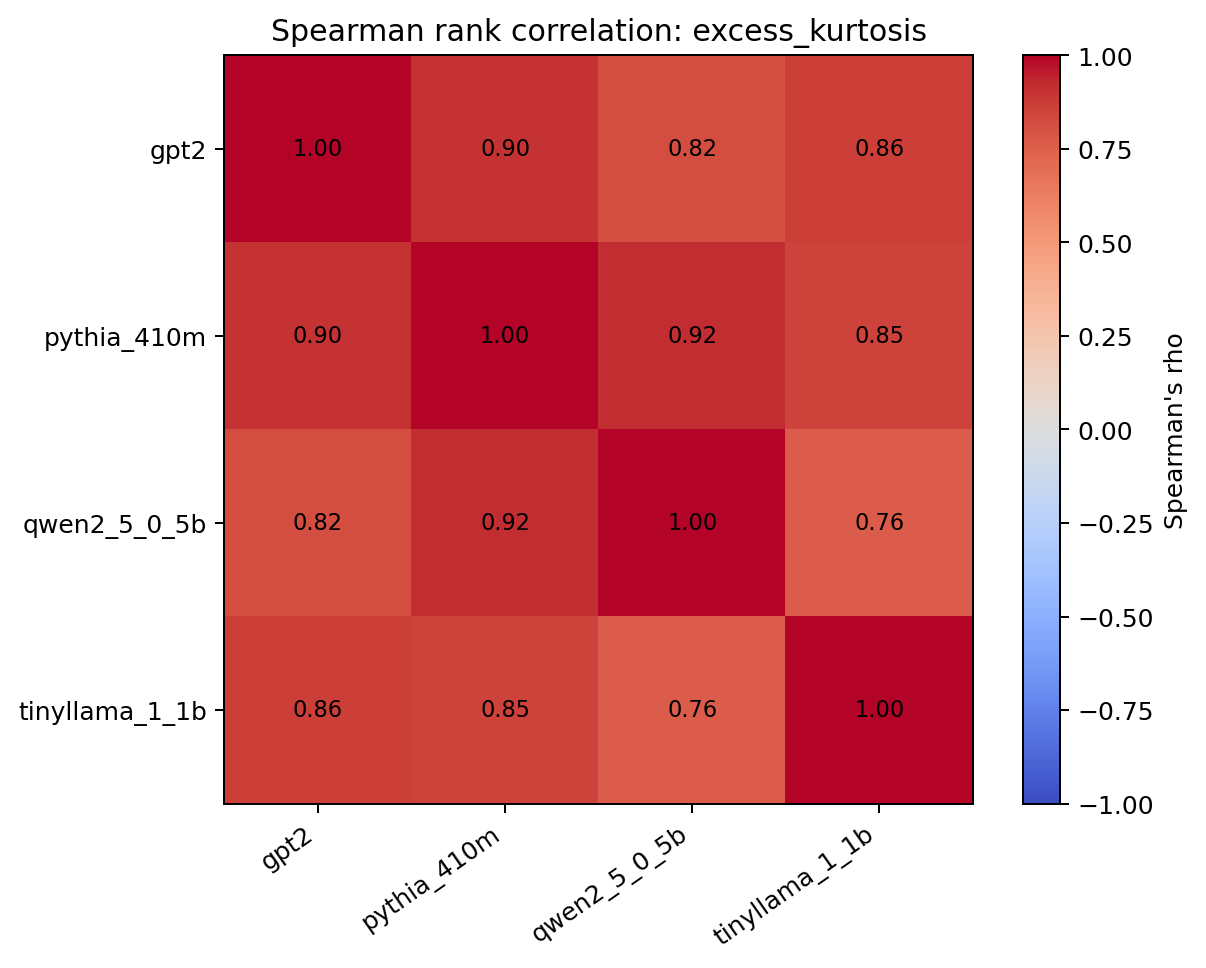}
        \subcaption{Excess Kurtosis ($\kappa-3$)}
        \label{fig:heat_kurt}
    \end{minipage}
\end{minipage}
\caption{Pairwise Spearman's rank correlation ($\rho$) of cross-entropy moments across the four evaluated language models for the poetic regime. The near-perfect agreement for mean and variance demonstrates the invariance of the metric, demonstrating that the metric survives across different model paradigms.}
\label{fig:poetic_heatmaps}
\end{figure*}

\begin{figure*}[h!]
\centering
\begin{minipage}{0.77\textwidth}
    \centering
    % Row 1: Mean and Variance
    \begin{minipage}{0.48\textwidth}
        \centering
        \includegraphics[width=\linewidth]{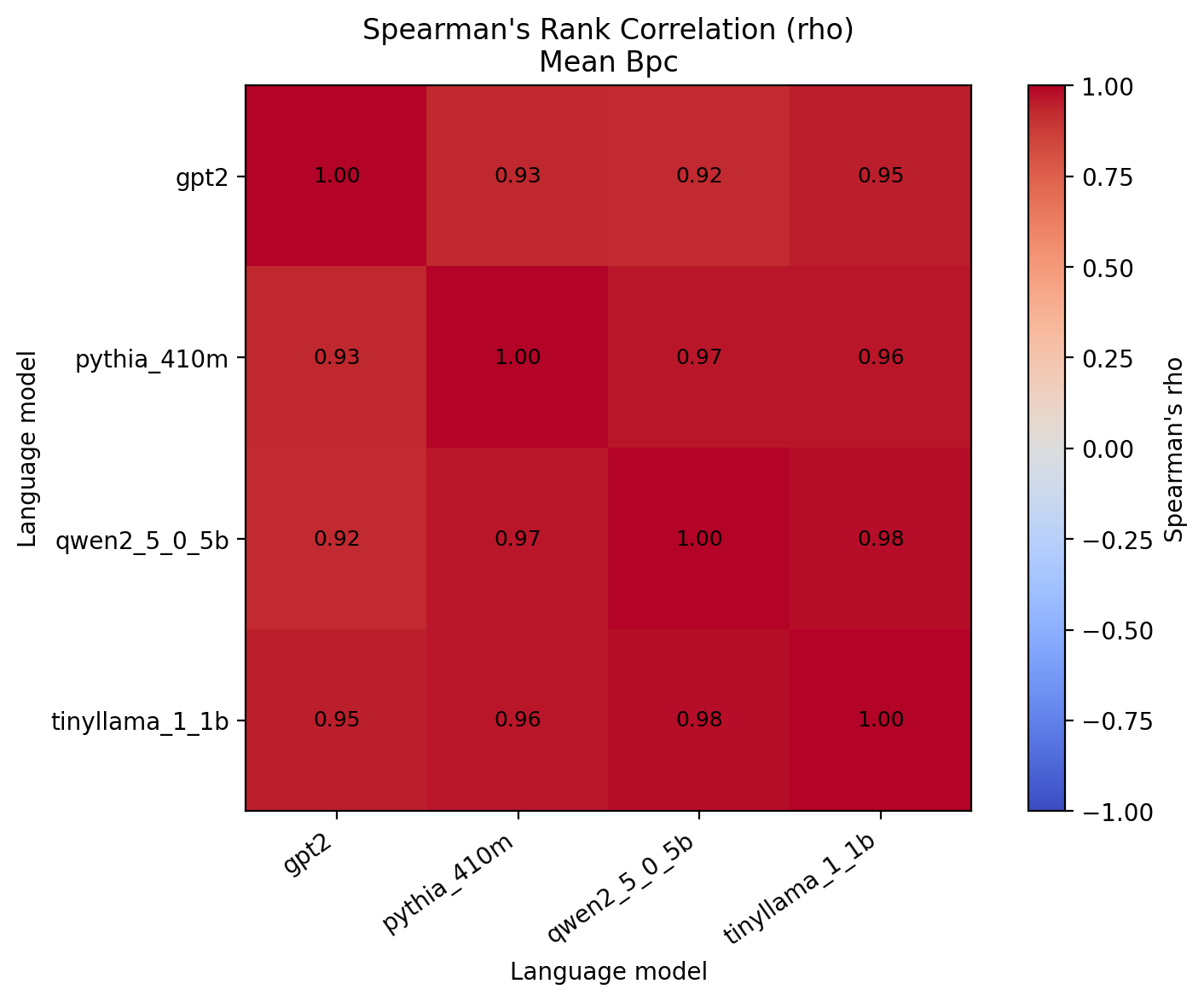}
        \subcaption{Mean ($\mu$)}
        \label{fig:acad_heat_mean}
    \end{minipage}\hfill
    \begin{minipage}{0.48\textwidth}
        \centering
        \includegraphics[width=\linewidth]{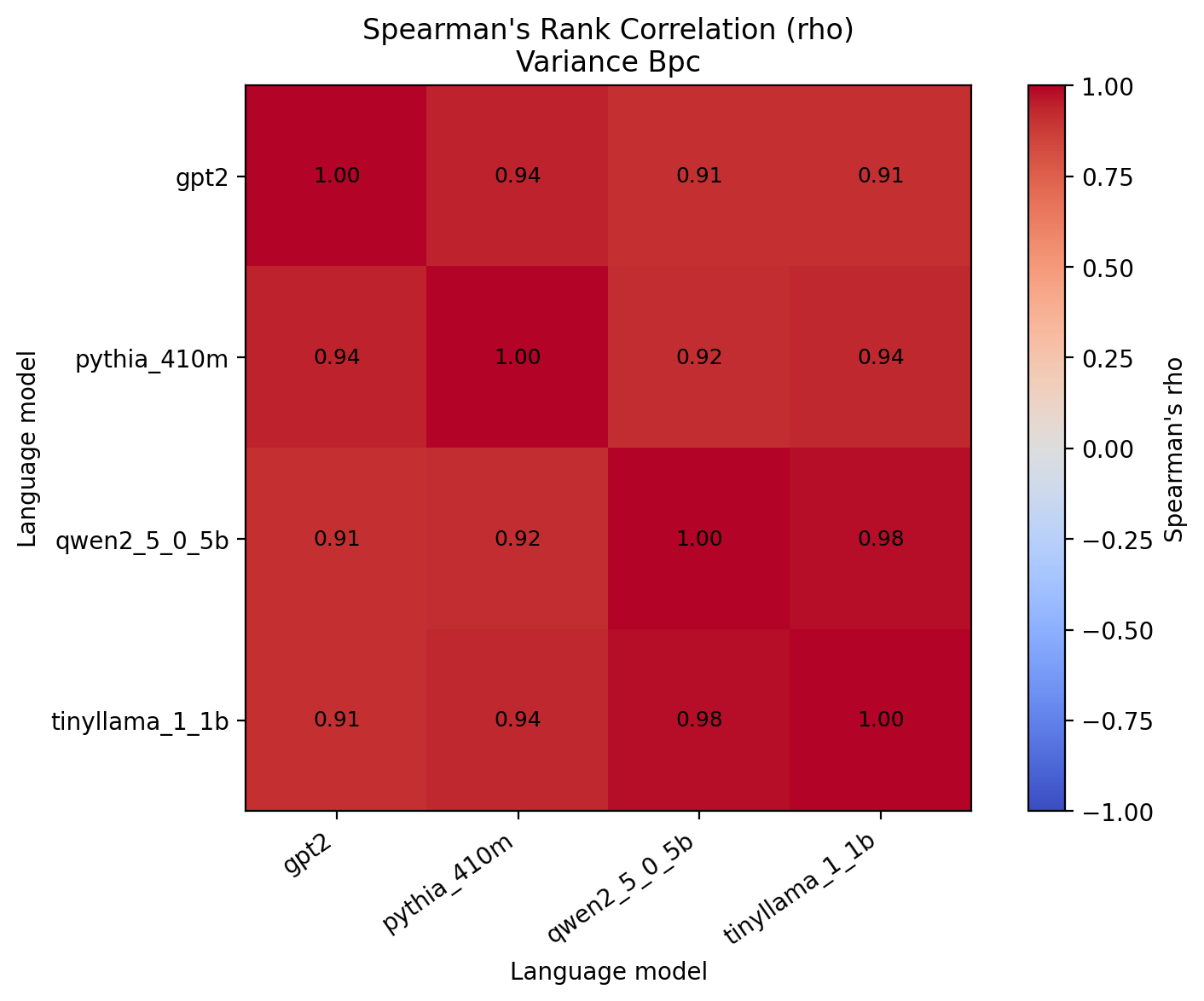}
        \subcaption{Variance ($\sigma^2$)}
        \label{fig:acad_heat_var}
    \end{minipage}

    \vspace{0.4em} 

    % Row 2: Skewness and Excess Kurtosis
    \begin{minipage}{0.48\textwidth}
        \centering
        \includegraphics[width=\linewidth]{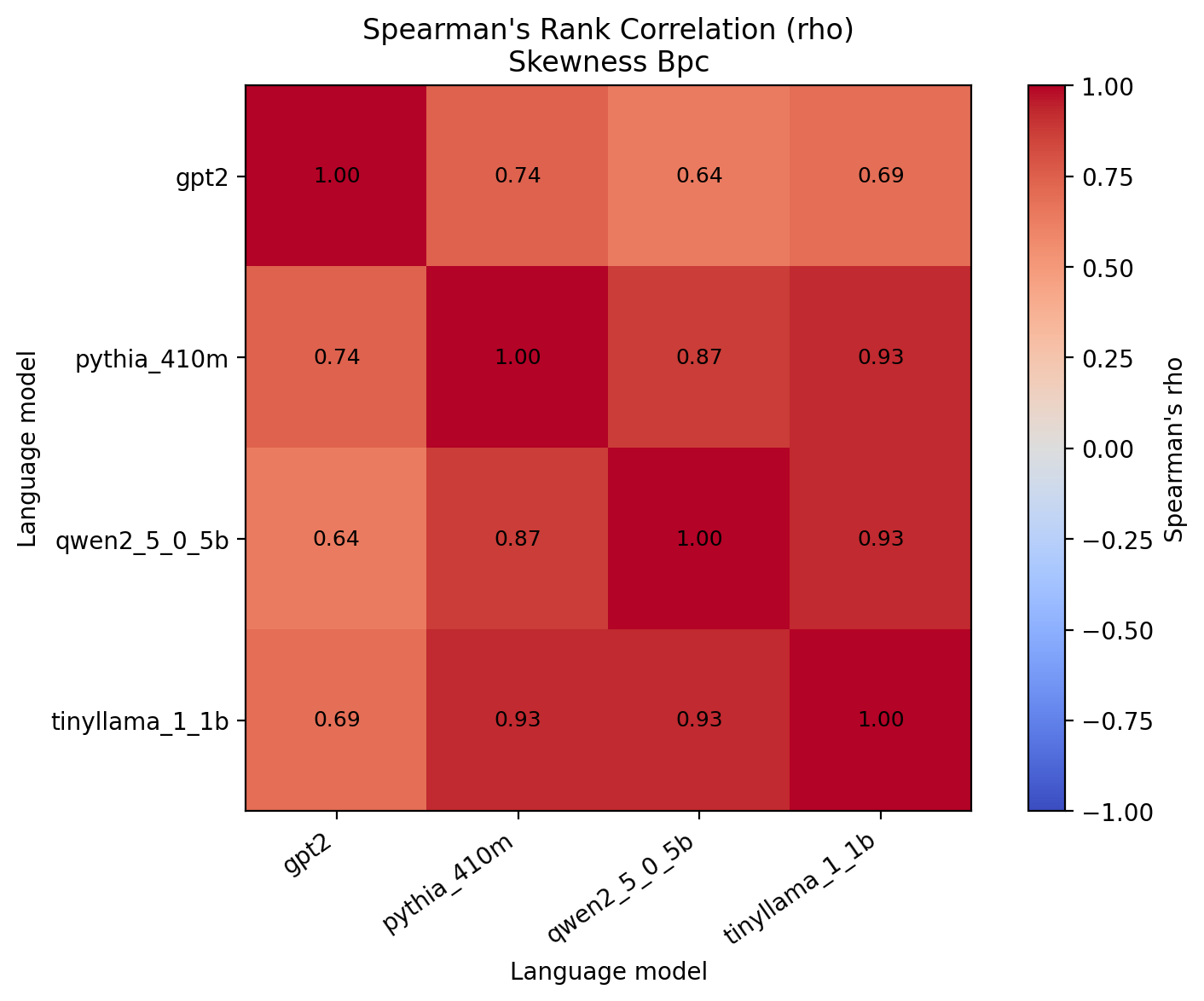}
        \subcaption{Skewness ($\gamma$)}
        \label{fig:acad_heat_skew}
    \end{minipage}\hfill
    \begin{minipage}{0.48\textwidth}
        \centering
        \includegraphics[width=\linewidth]{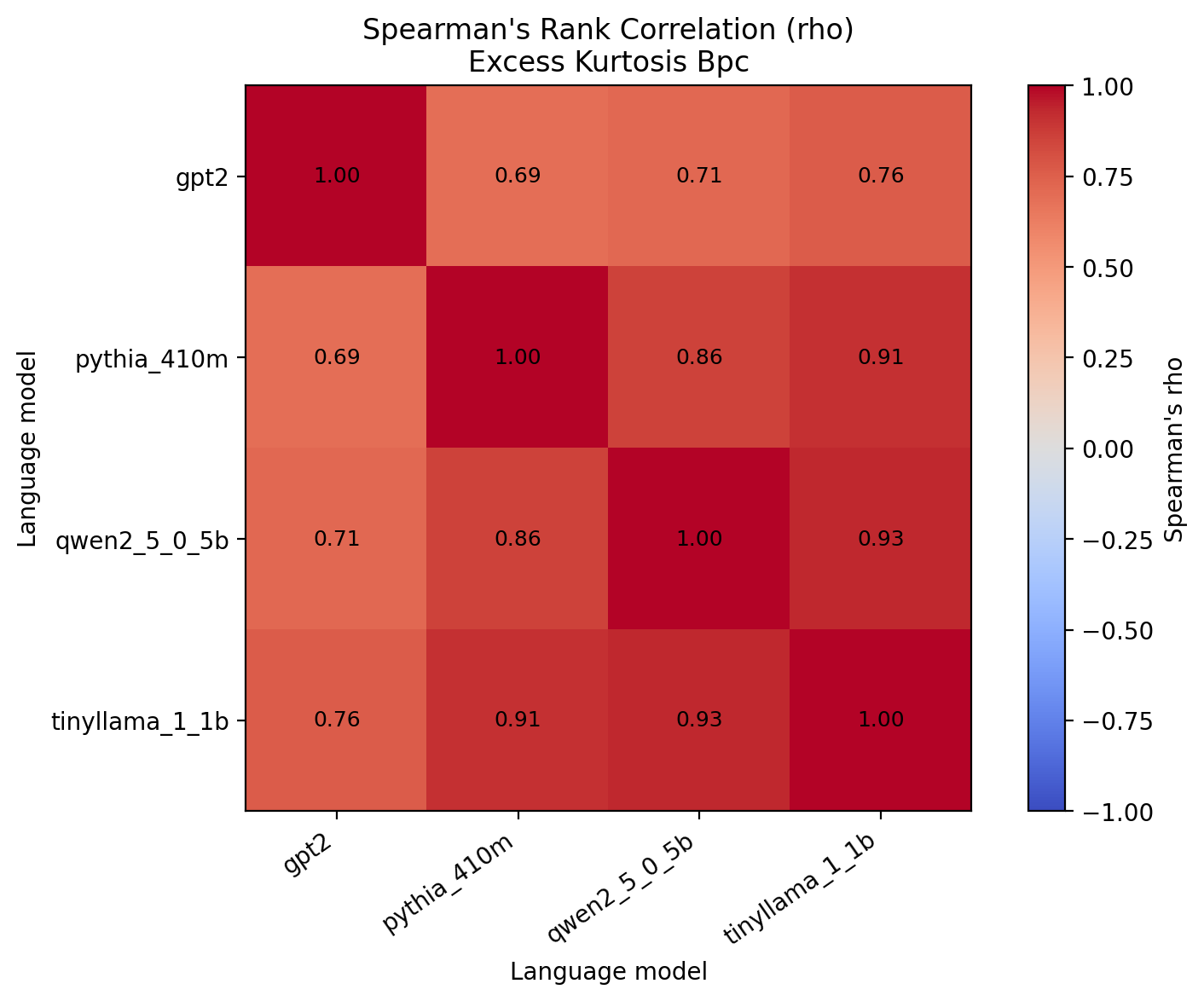}
        \subcaption{Excess Kurtosis ($\kappa-3$)}
        \label{fig:acad_heat_kurt}
    \end{minipage}
\end{minipage}
\caption{Pairwise Spearman's rank correlation ($\rho$) of cross-entropy moments across the four evaluated language models for the academic regime. Despite the extreme stylistic variance compression enforced by institutional formatting, first two moments (mean and variance) maintain robust invariance.}
\label{fig:academic_heatmaps}
\end{figure*}

\subsection{The Poetic Regime}
\label{res:poet}

We first evaluate the stability within the poetic regime ($S_{\text{poet}}$), a dataset characterized by high stylistic volatility and creative freedom. Despite the fundamental differences in vocabulary capacities and tokenization boundaries across \texttt{GPT-2}, \texttt{Pythia}, \texttt{TinyLlama}, and \texttt{Qwen}, the models exhibit near-perfect agreement on the relative ranking of the 28 poets.

For the baseline textual unpredictability, quantified by the mean cross-entropy ($\mu$), the pairwise Spearman rank correlation ($\rho$) across all model combinations ranges from $0.968$ to $0.994$. The structural volatility, captured by the variance ($\sigma^2$), demonstrates similarly robust invariance, with $\rho$ bounded strictly between $0.896$ and $0.988$. These near-ceiling correlations prove that the core stylistic fingerprint of a poet translates almost flawlessly across distinct model architectures. The confidence intervals are given in appendix \ref{app:bootstrap1}.

The stricter Kendall's rank correlation ($\tau$) corroborates this finding, maintaining values of $\tau > 0.878$ for the mean and $\tau > 0.846$ for the variance across all model pairs. 

As expected from statistical theory \cite{kendall1977advanced}, the higher-order moments — skewness ($\gamma$) and kurtosis ($\kappa$) — exhibit slightly wider variance ($\rho \in [0.689, 0.948]$ for skewness and $\rho \in [0.759, 0.915]$ for kurtosis). Because these moments are mathematically hypersensitive to extreme outliers in the chunk frequency distributions, minor subword segmentation differences at the model level naturally introduce localized noise. However, the correlations remain strongly positive.

These results establish our baseline: when human writing exhibits broad stylistic diversity, the relative statistical moments of language model surprisal are largely model-independent. This demonstrates that when utilizing character-normalized surprisal, the underlying stylistic ranking is not disrupted by the disparate tokenization boundaries of the models.

\subsection{The Academic Regime}
\label{res:acad}

We next evaluate the results within the academic regime ($S_{\text{acad}}$). Unlike contemporary poetry, published research papers adhere to rigid institutional formatting, standardized technical terminologies, and homogenous syntactic structures. This extreme structural constraint compresses the baseline stylistic variance across the corpus, narrowing the margins of difference between authors.

Despite this intense compression, the multi-model rank correlation remains remarkably robust. For the mean cross-entropy ($\mu$), the pairwise Spearman correlation ($\rho$) across \texttt{GPT-2}, \texttt{Pythia}, \texttt{TinyLlama}, and \texttt{Qwen} remains tightly bounded between $0.922$ and $0.978$. The structural volatility, measured by variance ($\sigma^2$), exhibits similarly high stability, with $\rho \in [0.907, 0.981]$. Kendall's $\tau$ corroborates this strong monotonic relationship, maintaining values between $0.759$ and $0.897$ for the first two moments.

Because the stylistic distances between academic papers are microscopically thin compared to poetry, we observe a wider fluctuation in the higher-order moments. Skewness ($\gamma$) yields $\rho \in [0.637, 0.929]$, while excess kurtosis ($\kappa-3$) ranges from $0.692$ to $0.934$. When attempting to rank highly clustered data points, the inherent noise of model-specific segmentation (e.g., how differently \texttt{Pythia} and \texttt{Qwen} split \LaTeX math symbols or citations) naturally causes minor rank swapping in the extreme tails of the frequency distribution. 

However, the fact that the mean and variance correlations comfortably exceed $0.90$ under such constrained conditions is highly significant (confidence intervals are given in appendix \ref{app:bootstrap2}). It implies that the extent of stylometric fingerprint of a text survives even when institutional rules force authors into a highly constrained structural space.

\paragraph{Localized Surprisal Spikes in \texttt{GPT-2}:} A notable pattern emerges in the higher-order moments: \texttt{GPT-2} acts as a distinct outlier relative to the other three architectures. While \texttt{Pythia}, \texttt{Qwen}, and \texttt{TinyLlama} exhibit strong mutual rank agreement for both skewness ($\rho \in [0.87, 0.93]$) and excess kurtosis ($\rho \in [0.86, 0.93]$), \texttt{GPT-2}'s correlation with these modern models degrades substantially ($\rho \in [0.64, 0.74]$ for skewness and $\rho \in [0.69, 0.76]$ for excess kurtosis). This disparity is significant and it suggests that \texttt{GPT-2} experiences \textit{localized surprisal spikes} during subword fragmentation, causing systematic rank shifts in skewness and kurtosis that are likely absent in the rest of the LLMs. 

Mechanistically, this disparity is likely driven by \texttt{GPT-2}’s older, limited tokenizer and its informal pre-training corpus (detailed in appendix \ref{app:model}). When processing dense academic artifacts such as mathematical equations and inline citations, \texttt{GPT-2} is prone to fragment them into highly unpredictable subword sequences, creating localized surprisal spikes. Because higher-order moments are mathematically hypersensitive to extreme outliers, these tokenizer failures disproportionately distort GPT-2’s skewness and kurtosis rankings within the tightly compressed academic regime.

\subsection{Compression Baseline}
\label{sec:compression-baseline}

To test whether high cross-model agreement merely reflects large differences in intrinsic complexity between sources, we replace the language models with model-free compression baselines. For each chunk, we compute compressed size in bits per character, using deflate, LZMA, bzip2, and zstd (\emph{plain}), as well as deflate and LZMA primed with a neutral English reference text disjoint from both corpora (\emph{primed}). Chunking and source-level moments are identical to those used for the language models. We compare the mean Spearman correlation among language-model pairs with the mean correlation between each compressor and the language models, and estimate their difference with a paired bootstrap over sources (full results in appendix~\ref{app:compression-baseline}).

The results of this comparison indicate that the language models agree with each other in several of the rankings where there is no concrete agreement between the baselines and the models. This suggest that all four models, despite their differences, simultaneously capture extents of randomness which are broader than generic textual differences captured by the baselines. However the gap is narrowed for rankings based on the third and fourth moments.  

\subsubsection{Poetic Regime}

Plain compressors agree with the language models far less than the language models agree with each other on all four moments, so we focus on the primed variants. For the mean, primed compressors remain clearly below language-model agreement. For the variance, the gap is smaller: primed LZMA nearly matches the language models on some cases, suggesting that variance rankings may reflect properties of the texts that a compressor, though unreliable, may reproduce. For skewness and excess kurtosis, primed compressors are in most cases statistically indistinguishable from the language models, so we find no evidence that these moments carry language-model-specific signal. 

\subsubsection{Academic Regime}

The pattern differs from that in poetry. Plain compressors again agree with the language models far less than the language models agree with each other, and the primed variants are closer but still clearly below on every moment. For the mean, the best primed compressor (LZMA, $\bar\rho=0.78$) remains well below language-model agreement ($\bar\rho=0.97$), and no compressor reaches the threshold fixed beforehand. For the variance, the gap is substantially larger than in the poetic corpus, so here the stability of variance rankings is not reproduced by a model-free measure. Unlike in poetry, skewness and excess kurtosis also separate clearly: primed compressors reach only $\bar\rho\approx0.3$--$0.5$, against roughly $0.9$ among the language models, and every bootstrap interval for the difference excludes zero. One possible reading is that language models agree with each other, and compressors do not, precisely where inter-source differences are small. Because the primed reference is neutral English rather than technical text, the baseline may understate what a compressor adapted to academic prose could achieve (full results in Appendix~\ref{app:compression-baseline-academic}).

\subsection{The Geometry of Stability}

We now use \texttt{GPT-2} cross-entropy values to analyze the geometrical positions of the poets and the papers in the mean-variance phase space.

\subsubsection{Spatial Dispersion in the Poetic Regime}

To understand the mechanical origin of this cross-model rank stability, we project the textual predictability of each corpus into a two-dimensional moment phase space defined by the mean ($\mu$) and variance ($\sigma^2$) of the cross-entropy.

\begin{figure*}[h!]
\centering
\begin{minipage}{0.75\textwidth} % This outer minipage compresses the total footprint
    \centering
    % Figure A: Poetic Regime
    \begin{minipage}[b]{0.48\textwidth}
        \centering
        \includegraphics[width=\linewidth]{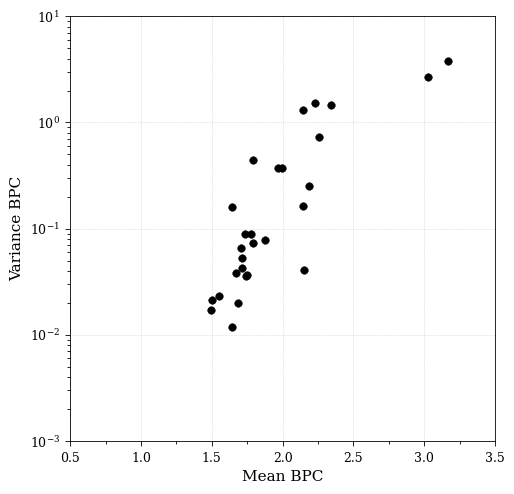}
        \caption{Mean and variance of source-level \texttt{GPT-2} cross-entropy for the poetry corpus.}
        \label{fig:poet_mean_variance}
    \end{minipage}\hfill
    % Figure B: Academic Regime
    \begin{minipage}[b]{0.48\textwidth}
        \centering
        \includegraphics[width=\linewidth]{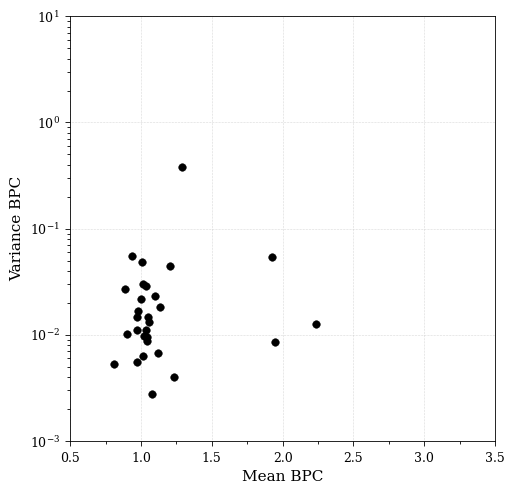}
        \caption{Mean and variance of source-level \texttt{GPT-2} cross-entropy for the research-paper corpus.}
        \label{fig:paper_mean_variance}
    \end{minipage}
\end{minipage}
\end{figure*}

Evaluating the poetic regime ($S_{\text{poet}}$) under \texttt{GPT-2} reveals a massive spatial footprint. Because poets intentionally manipulate syntax, rhythmic density, vocabulary and other stylistic measures, their individual surprisal signatures occupy more or less distinct, separated regions of the moment space, as shown in figure \ref{fig:poet_mean_variance}. Across the 28 evaluated poets, the mean unpredictability spans a wide domain ($\mu \in [1.49, 3.17]$), while the variance spans several orders of magnitude ($\sigma^2 \in [0.01, 3.76]$).

This wide dispersion may have been the mathematical driver of the near-perfect rank correlation observed in Section \ref{res:poet}. When the inter-author distances ($\Delta \mu, \Delta \sigma^2$) are vast relative to the minor perturbations introduced by different tokenization paradigms ($\epsilon$), the relative rank ordering becomes virtually immune to architectural noise. The separation of coordinates in the unconstrained regime guarantees that model-specific biases cannot shift an author's statistical fingerprint enough to alter their macroscopic rank within the corpus. However, some of the points in this plot are close as well, and these prevent a correlation of exactly unity.

\subsubsection{Compression and Clustered Noise in the Academic Regime}

Conversely, evaluating the academic corpus ($S_{\text{acad}}$) reveals a severe geometric contraction in the moment phase space, as shown is figure \ref{fig:paper_mean_variance}. Driven by institutional formatting, standardized technical vocabularies, and rigid syntactic structures, the 29 research papers collapse into a highly dense, localized cluster. Under \texttt{GPT-2}, the baseline unpredictability for the academic texts drops significantly ($\mu \in [0.80, 2.23]$), with the vast majority of the corpus (75th percentile) falling below $1.12$. More strikingly, the stylistic volatility is almost entirely eliminated, with variance heavily compressed ($\sigma^2 \in [0.002, 0.382]$) and a median variance of just $0.013$.

This macroscopic compression mechanically explains the behavior of the rank correlations observed in Section \ref{res:acad}. In this densely packed regime, the underlying stylistic differences between papers may become microscopically small ($\Delta \mu \approx \epsilon$). When data points are clustered this tightly, localized tokenization artifacts—such as how a specific model segments \LaTeX mathematical equations, embedded citations, or compound technical nouns—become large enough relative to the inter-author distances to perturb the local rank ordering. 

Because higher-order moments ($\gamma, \kappa$) are mathematically hypersensitive to sudden, isolated surprisal spikes, these minor tokenization variations cause localized rank-swapping in the distribution tails. However, the foundational statistical mass remains stable enough that the lower-order moments ($\mu, \sigma^2$) absorb the noise, preserving their $> 0.90$ rank correlations. This contrast indicates that the stylometric fingerprint remains model-agnostic even in the case of strict structural bottlenecks.

\subsubsection{Correlation between Mean and Variance of Cross-Entropy}
There is also another interesting result that should be looked into closely. There is a highly positive correlation between the means and variances of the cross-entropy of the 28 poets, in the corpus of the poetries. We observe that Pearson's correlation coefficient, $r=0.907$, which is found to be statistically significant ($p=3 \times 10^{-11}$). This means that poets whose writings were more unpredictable on average, were also more widespread across their values of cross-entropy of their chunks. This means that different chunks of their poetries surprised the model to different degrees, leading to a wide range of cross-entropy.

However, for the set of research papers, there was no meaningful correlation and Pearson's correlation coefficient was calculated to be $r=0.111$, and this is found to be statistically insignificant ($p=0.565$). This is also visible in figure \ref{fig:paper_mean_variance}, as the cluster formed by most of the points, is smooth. This indicates that mean and variance were almost orthogonal to each other, i.e., there was no meaningful correlation. 

\section{Conclusion}
In this work, we showed an empirical evidence that cross-entropy measures of literary and academic works are largely independent of the language model it is computed with respect to. One way of showing this would have been to simply choose some textual chunks and rank the cross-entropy measures of different models of those random chunks. In our work, we show that the rankings of these measures are stable even when the measures are considered across different poets and different published articles. Therefore, the phenomenon is shown to be robust across completely different sources (poets / research articles) of two different genres of writing. Further, the fact that the chosen language models differ inherently, indicates that this is not an obvious outcome. As evident from our findings, relative extents of textual unpredictability, when they are measured by the cross-entropy with respect to a language model, may be independent of the language model. This means that, for a given dataset, the rank of its elements based on cross-entropy measures may be completely independent of the model used.

\section*{Limitations}

While this study establishes the robust cross-model rank correlation of relative cross-entropy measures, its scope is bounded by three primary constraints:

\begin{itemize}
    \item \textbf{Model Scale:} The evaluation is restricted to four relatively small-scale language models. Future work has to evaluate massive-scale architectures to determine if increased parameter counts further smooth the localized surprisal spikes observed in higher-order statistical moments.
    \item \textbf{Stylistic Extremes:} By analyzing only two highly contrasting genres --- unconstrained contemporary poetry and rigidly structured academic articles --- the results leave open the behavior of intermediate text types, such as narrative fiction or journalistic prose.
    \item \textbf{Monolingual Focus:} The current corpora are entirely English-language. Broadening this scope to encompass multilingual datasets is necessary to understand how distinct morphological structures alter tokenization dynamics.
\end{itemize}

\section*{Ethics Statement}
This research analyzes contemporary poetry and published academic manuscripts that may be subject to copyright. Their analysis within this study was conducted strictly under fair use provisions for non-commercial academic research. The details of the poets and papers analyzed in this work, and the entire codebase are available on the GitHub repository of the project, linked on the first page. 

\bibliography{references}

\appendix

\section{Construction of a Chunk}
\label{app:chunk}
The chunk construction mechanism is similar for both corpora. However, in the corpus of poets, we consider the set of all available poems written by the poet. In the corpus of research papers, we consider the content of the whole paper except some specific parts. The chunk construction for both cases are described in the two following subsections. 

\subsection{Chunking Methodology for the Poetic Corpus}
For the unconstrained poetic regime, text segmentation is strictly stanza-preserving to respect the structural intent of the author. The chunking algorithm adheres to the following rules:
\begin{itemize}
    \item A chunk accumulates complete, contiguous stanzas until it reaches a minimum threshold of 200 words, i.e., it has $N$ number of stanzas such that the total word count of $N-1$ stanzas is less than 200 but that of $N$ stanzas is not less than 200.
    \item Stanzas are never mathematically or textually split across multiple chunks.
    \item If an entire poem contains fewer than 200 words, it is retained intact as a single chunk.
    \item For longer poems, if the final accumulated chunk falls below the 200-word threshold, it is merged into the preceding chunk to avoid anomalous undersized remainders.
\end{itemize}

Crucially, the empirical frequency distribution of cross-entropies is constructed at the \textbf{poet-level}, not the poem-level. All valid chunks extracted across a specific poet's entire evaluated collection are binned into a single, unified frequency histogram (using Freedman–Diaconis binning \cite{freedman1981histogram} capped between 8 and 30 bins). The statistical moments (mean, variance, skewness, kurtosis) are derived from this aggregate poet-level distribution.

\subsection{Chunking Methodology for the Academic Corpus}
For the constrained academic regime, segmentation respects logical manuscript boundaries by being strictly subsection-local and paragraph-preserving. The extraction follows these parameters:

\begin{itemize}
    \item Complete paragraphs are accumulated sequentially until the chunk reaches a minimum of 200 words, i.e., it has $N$ number of paragraphs such that the total word count of $N-1$ paragraphs is less than 200 but that of $N$ paragraphs is not less than 200.
    \item Paragraphs are never split, and chunks are strictly forbidden from crossing subsection boundaries.
    \item If a discrete subsection contains fewer than 200 words in its entirety, it is isolated as a single unified chunk.
    \item If the final paragraphs of a subsection form a remainder below the 200-word threshold, they are appended to the immediately preceding chunk within that same subsection.
\end{itemize}

Unlike the poetic corpus, the empirical frequency distribution for this regime is constructed at the \textbf{paper-level}. All chunks extracted from a single research manuscript are binned into a dedicated frequency histogram. The structural moments are then computed from this paper-specific distribution to represent the stylometric fingerprint of the individual manuscript.

\section{Mathematical Formulations of Moments and Rank Correlation}
\label{app:moment}

This section details the formal mathematical extraction of the statistical moments and the rank correlation metric discussed in Section 3. Let $h_{i,j} = H^{(M_k)}(c_{i,j})$ denote the cross-entropy of the $j$-th chunk from source $i$ under language model $M_k$. 

The \textbf{Mean} ($\mu_i$) establishes the baseline textual unpredictability of the source:
\begin{equation}
\mu_i = \frac{1}{K_i} \sum_{j=1}^{K_i} h_{i,j}.
\end{equation}

The \textbf{Variance} ($\sigma_i^2$) measures the structural volatility, quantifying how widely the stylistic complexity fluctuates across the text:
\begin{equation}
\sigma_i^2 = \frac{1}{K_i} \sum_{j=1}^{K_i} (h_{i,j} - \mu_i)^2.
\end{equation}

The \textbf{Skewness} ($\gamma_i$) evaluates the asymmetry of the distribution, identifying whether a source leans toward infrequent bursts of highly unexpected text or prolonged spans of highly predictable text:
\begin{equation}
\gamma_i = \frac{1}{K_i} \sum_{j=1}^{K_i} \left( \frac{h_{i,j} - \mu_i}{\sigma_i} \right)^3.
\end{equation}

The \textbf{Kurtosis} ($\kappa_i$) measures the presence of extreme structural deviations, capturing the frequency of highly anomalous chunks relative to the author's baseline style:
\begin{equation}
\kappa_i = \frac{1}{K_i} \sum_{j=1}^{K_i} \left( \frac{h_{i,j} - \mu_i}{\sigma_i} \right)^4.
\end{equation}

To quantify the ranking-agreement between any pair of models $M_k, M_l \in \mathcal{M}$, we compute Spearman's rank correlation coefficient:
\begin{equation}
\rho = 1 - \frac{6 \sum_{i=1}^{|S|} d_i^2}{|S|(|S|^2 - 1)},
\end{equation}
where $d_i = R_i^{(M_k)} - R_i^{(M_l)}$ is the difference in the assigned ranks for source $i$, and $|S|$ is the total number of sources in the respective corpus ($|S_{\text{poet}}| = 28$ or $|S_{\text{acad}}| = 29$).

In addition to Spearman's rank correlation, we compute Kendall's rank correlation coefficient ($\tau$) to further validate the monotonic agreement between the models. For any pair of rank vectors $R^{(M_k)}$ and $R^{(M_l)}$, Kendall's $\tau$ is defined as:
\begin{equation}
\tau = \frac{2(n_c - n_d)}{|S|(|S| - 1)},
\end{equation}
where $n_c$ represents the number of concordant pairs, $n_d$ represents the number of discordant pairs, and the denominator denotes the total number of possible pair combinations among the $\vert{}S\vert{}$ sources.

\begin{table*}[h!]
\centering
\small
\setlength{\tabcolsep}{5pt}
\begin{tabular}{lcccc}
\hline
\textbf{Model pair} & \textbf{Mean BPC} & \textbf{Variance} & \textbf{Skewness} & \textbf{Excess kurtosis} \\
\hline
\texttt{GPT-2}--\texttt{Pythia} & [0.919, 0.995] & [0.814, 0.996] & [0.778, 0.985] & [0.775, 0.959] \\
\texttt{GPT-2}--\texttt{Qwen} & [0.914, 0.993] & [0.787, 0.993] & [0.464, 0.947] & [0.588, 0.926] \\
\texttt{GPT-2}--\texttt{TinyLlama} & [0.898, 0.992] & [0.923, 0.994] & [0.832, 0.992] & [0.646, 0.948] \\
\texttt{Pythia}--\texttt{Qwen} & [0.974, 0.999] & [0.945, 0.998] & [0.663, 0.970] & [0.762, 0.975] \\
\texttt{Pythia}--\texttt{TinyLlama} & [0.922, 0.999] & [0.684, 0.996] & [0.691, 0.968] & [0.641, 0.951] \\
\texttt{Qwen}--\texttt{TinyLlama} & [0.897, 1.000] & [0.655, 0.997] & [0.318, 0.913] & [0.458, 0.935] \\
\hline
\end{tabular}
\caption{95\% paired-bootstrap confidence intervals for Spearman's rank correlation coefficients across the four measures in the poetic corpus ($n=28$ poets).}
\label{tab:poetry_bootstrap_ci}
\end{table*}

\begin{table*}[h!]
\centering
\small
\setlength{\tabcolsep}{5pt}
\begin{tabular}{lcccc}
\hline
\textbf{Model pair} & \textbf{Mean BPC} & \textbf{Variance} & \textbf{Skewness} & \textbf{Excess kurtosis} \\
\hline
\texttt{GPT-2}--\texttt{Pythia} & [0.803, 0.989] & [0.812, 0.982] & [0.460, 0.914] & [0.407, 0.855] \\
\texttt{GPT-2}--\texttt{Qwen} & [0.772, 0.986] & [0.776, 0.969] & [0.234, 0.916] & [0.462, 0.857] \\
\texttt{GPT-2}--\texttt{TinyLlama} & [0.846, 0.984] & [0.765, 0.963] & [0.349, 0.906] & [0.528, 0.888] \\
\texttt{Pythia}--\texttt{Qwen} & [0.882, 0.991] & [0.784, 0.974] & [0.696, 0.947] & [0.664, 0.945] \\
\texttt{Pythia}--\texttt{TinyLlama} & [0.880, 0.989] & [0.824, 0.973] & [0.791, 0.977] & [0.790, 0.962] \\
\texttt{Qwen}--\texttt{TinyLlama} & [0.927, 0.994] & [0.934, 0.993] & [0.803, 0.978] & [0.820, 0.976] \\
\hline
\end{tabular}
\caption{95\% paired-bootstrap confidence intervals for Spearman's rank correlation coefficients across the four measures in the academic corpus ($n=29$ papers).}
\label{tab:academic_bootstrap_ci}
\end{table*}

\section{Bootstrap Confidence Intervals for Cross-Model Rank Correlations}
\label{app:bootstrap}

We estimate 95\% confidence intervals for Spearman's rank correlation coefficients using 5,000 paired-bootstrap resamples \cite{dror2018hitchhikers} of the sources. The intervals quantify uncertainty in the estimated cross-model rank correlations.

\subsection{Poetic Corpus}
\label{app:bootstrap1}
Table~\ref{tab:poetry_bootstrap_ci} reports the pairwise correlations and confidence intervals for the poetic corpus. Mean BPC exhibits consistently high rank agreement across models. Variance also shows strong agreement, although some confidence intervals are relatively wide. Skewness and excess kurtosis exhibit greater variability in both their correlations and confidence intervals, which indicate that these are not as robust as those for mean and variance. All intervals remain above zero, indicating positive rank associations across the evaluated model pairs.

\subsection{Academic Corpus}
\label{app:bootstrap2}
Table~\ref{tab:academic_bootstrap_ci} presents the corresponding results for the academic corpus. Mean BPC and variance show consistently strong rank agreement across the evaluated models. In comparison, similar to the poetic corpus, skewness and excess kurtosis exhibit weaker correlations for some model pairs and wider confidence intervals. Nevertheless, all confidence intervals remain above zero, indicating positive cross-model rank associations across the four measures.

\section{Compression Baseline}
Here we discuss the numerical results obtained during the comparative study with the baseline.

\subsection{Results in Poetic Regime}
\label{app:compression-baseline}
\paragraph{Setup:}
Plain variants compress each chunk independently (deflate, LZMA, bzip2, zstd). Primed variants (deflate, LZMA) compress a neutral English reference
text $R$ concatenated with the chunk, and the chunk's cost is the size of $R$+chunk minus the size of $R$ alone. % TODO: state what $R$ is (source, size).
Bits per character use character length, as for the language models. For each moment we report (i) $\bar\rho_{\mathrm{LM\text{-}LM}}$, the mean
Spearman correlation over the six language-model pairs; (ii) $\bar\rho_{\mathrm{comp\text{-}LM}}$, the mean correlation between the compressor and the four language models; and (iii) $\Delta=\bar\rho_{\mathrm{LM\text{-}LM}}-\bar\rho_{\mathrm{comp\text{-}LM}}$ with a 95\% paired-bootstrap interval (5{,}000 resamples of sources; all
resamples valid). Intervals containing zero are marked with $\dagger$. The correlation coefficients and confidence intervals for this dataset are given in table~\ref{tab:baseline-binned}.

\begin{table}[h!]
\centering
\small
\resizebox{\columnwidth}{!}{%
\begin{tabular}{llrrc}
\hline
\textbf{Moment} & \textbf{Compressor} & $\bar\rho_{\mathrm{comp\text{-}LM}}$ & $\Delta$ & \textbf{95\% CI of $\Delta$} \\
\hline
\multicolumn{5}{l}{\emph{Mean} ($\bar\rho_{\mathrm{LM\text{-}LM}}=0.963$, $n=28$)} \\
 & plain deflate  & 0.505 & 0.458 & [0.156, 0.806] \\
 & plain LZMA     & 0.520 & 0.443 & [0.166, 0.781] \\
 & plain bzip2    & 0.580 & 0.383 & [0.184, 0.661] \\
 & plain zstd     & 0.579 & 0.384 & [0.170, 0.677] \\
 & primed deflate & 0.768 & 0.195 & [0.056, 0.431] \\
 & primed LZMA    & 0.805 & 0.158 & [0.047, 0.371] \\
\hline
\multicolumn{5}{l}{\emph{Variance} ($\bar\rho_{\mathrm{LM\text{-}LM}}=0.957$, $n=28$)} \\
 & plain deflate  & 0.803 & 0.154 & [0.060, 0.330] \\
 & plain LZMA     & 0.811 & 0.146 & [0.063, 0.304] \\
 & plain bzip2    & 0.759 & 0.198 & [0.087, 0.401] \\
 & plain zstd     & 0.786 & 0.171 & [0.072, 0.355] \\
 & primed deflate & 0.887 & 0.070 & [0.016, 0.196] \\
 & primed LZMA    & 0.918 & 0.039 & [0.008, 0.109] \\
\hline
\multicolumn{5}{l}{\emph{Skewness} ($\bar\rho_{\mathrm{LM\text{-}LM}}=0.847$, $n=28$)} \\
 & plain deflate  & 0.589 & 0.258 & [0.047, 0.529] \\
 & plain LZMA     & 0.534 & 0.314 & [0.101, 0.598] \\
 & plain bzip2    & 0.468 & 0.380 & [0.122, 0.711] \\
 & plain zstd     & 0.550 & 0.298 & [0.068, 0.604] \\
 & primed deflate & 0.712 & 0.136 & [$-$0.009, 0.346]$^\dagger$ \\
 & primed LZMA    & 0.764 & 0.083 & [$-$0.049, 0.289]$^\dagger$ \\
\hline
\multicolumn{5}{l}{\emph{Excess kurtosis} ($\bar\rho_{\mathrm{LM\text{-}LM}}=0.851$, $n=28$)} \\
 & plain deflate  & 0.549 & 0.302 & [0.113, 0.548] \\
 & plain LZMA     & 0.413 & 0.438 & [0.148, 0.798] \\
 & plain bzip2    & 0.480 & 0.371 & [0.107, 0.691] \\
 & plain zstd     & 0.492 & 0.360 & [0.095, 0.706] \\
 & primed deflate & 0.736 & 0.115 & [0.004, 0.258] \\
 & primed LZMA    & 0.756 & 0.095 & [$-$0.005, 0.219]$^\dagger$ \\
\hline
\end{tabular}%
}
\caption{Compressor baselines versus language models, poetic corpus. $^\dagger$: interval contains zero.}
\label{tab:baseline-binned}
\end{table}

\subsection{Results in Academic Regime}
\label{app:compression-baseline-academic}

The setup is identical to appendix~\ref{app:compression-baseline}. We report moments computed from the binned chunk-level distribution, over $n=29$ papers for all four moments. $\bar\rho_{\mathrm{LM\text{-}LM}}$ is the mean Spearman correlation over the six language-model pairs, $\bar\rho_{\mathrm{comp\text{-}LM}}$ is the mean correlation between the compressor and the four language models, and $\Delta=\bar\rho_{\mathrm{LM\text{-}LM}}-\bar\rho_{\mathrm{comp\text{-}LM}}$ is reported with a 95\% paired-bootstrap interval (5{,}000 resamples of papers, all valid). All intervals exclude zero. It is to be noted that, specifically for this set, for an apples-to-apples comparison with the baseline, we slightly modified the processing of the papers. Due to that, the values of $\bar\rho_{\mathrm{LM\text{-}LM}}$ may not match exactly with the values reported in section \ref{sec:results}, although the remain close. The correlation coefficients and confidence intervals for this dataset are given in table~\ref{tab:baseline-academic}.

\begin{table}[h!]
\centering
\small
\resizebox{\columnwidth}{!}{%
\begin{tabular}{llrrc}
\hline
\textbf{Moment} & \textbf{Compressor} & $\bar\rho_{\mathrm{comp\text{-}LM}}$ & $\Delta$ & \textbf{95\% CI of $\Delta$} \\
\hline
\multicolumn{5}{l}{\emph{Mean} ($\bar\rho_{\mathrm{LM\text{-}LM}}=0.975$, $n=29$)} \\
 & plain deflate  & 0.463 & 0.512 & [0.194, 0.908] \\
 & plain LZMA     & 0.379 & 0.596 & [0.200, 1.019] \\
 & plain bzip2    & 0.355 & 0.620 & [0.269, 1.026] \\
 & plain zstd     & 0.363 & 0.612 & [0.260, 1.014] \\
 & primed deflate & 0.700 & 0.275 & [0.130, 0.508] \\
 & primed LZMA    & 0.779 & 0.196 & [0.084, 0.395] \\
\hline
\multicolumn{5}{l}{\emph{Variance} ($\bar\rho_{\mathrm{LM\text{-}LM}}=0.961$, $n=29$)} \\
 & plain deflate  & 0.553 & 0.408 & [0.153, 0.735] \\
 & plain LZMA     & 0.454 & 0.507 & [0.207, 0.884] \\
 & plain bzip2    & 0.415 & 0.545 & [0.240, 0.934] \\
 & plain zstd     & 0.553 & 0.408 & [0.148, 0.739] \\
 & primed deflate & 0.725 & 0.236 & [0.066, 0.485] \\
 & primed LZMA    & 0.704 & 0.257 & [0.073, 0.506] \\
\hline
\multicolumn{5}{l}{\emph{Skewness} ($\bar\rho_{\mathrm{LM\text{-}LM}}=0.909$, $n=29$)} \\
 & plain deflate  & 0.321 & 0.588 & [0.270, 0.956] \\
 & plain LZMA     & 0.213 & 0.696 & [0.380, 1.061] \\
 & plain bzip2    & 0.261 & 0.647 & [0.321, 1.028] \\
 & plain zstd     & 0.237 & 0.671 & [0.339, 1.057] \\
 & primed deflate & 0.473 & 0.436 & [0.195, 0.754] \\
 & primed LZMA    & 0.397 & 0.512 & [0.231, 0.846] \\
\hline
\multicolumn{5}{l}{\emph{Excess kurtosis} ($\bar\rho_{\mathrm{LM\text{-}LM}}=0.897$, $n=29$)} \\
 & plain deflate  & 0.211 & 0.685 & [0.358, 1.037] \\
 & plain LZMA     & 0.274 & 0.623 & [0.294, 0.985] \\
 & plain bzip2    & 0.285 & 0.612 & [0.282, 0.975] \\
 & plain zstd     & 0.278 & 0.619 & [0.277, 0.988] \\
 & primed deflate & 0.333 & 0.564 & [0.265, 0.945] \\
 & primed LZMA    & 0.300 & 0.597 & [0.267, 0.988] \\
\hline
\end{tabular}%
}
\caption{Compressor baselines versus language models, academic corpus.}
\label{tab:baseline-academic}
\end{table}

\section{Architectural and Training Disparities of Studied Models}
\label{app:model}

To provide context for the rank correlation results presented in Section \ref{sec:results}, we outline the structural and training differences of the four language models evaluated. Because these models span distinct generational architectures and training domains, their convergence on stylometric rankings is not an obvious consequence of shared design principles.

\paragraph{\texttt{GPT-2}:} 
\texttt{GPT-2} represents an early autoregressive Transformer decoder utilizing absolute positional embeddings. It employs a standard Byte-Pair Encoding (BPE) tokenizer with a vocabulary size of 50,257 subword units. The model was pre-trained on WebText, an uncurated dataset constructed from outbound Reddit links, which primarily consists of conversational internet English.

\paragraph{\texttt{Pythia}:} 
This model is based on the \texttt{GPT-NeoX} architecture. It incorporates Rotary Positional Embeddings (RoPE) \cite{su2024roformer} and computes attention and multilayer perceptron (MLP) blocks in parallel. It uses the NeoX BPE tokenizer with 50,304 subword units. \texttt{Pythia} was trained on The Pile, a highly curated dataset that includes a high density of academic text, biomedical literature, and formal prose.

\paragraph{\texttt{TinyLlama}:} 
\texttt{TinyLlama} adopts the Llama-2 architectural framework, featuring Grouped-Query Attention (GQA) \cite{ainslie2023gqa}, SwiGLU \cite{shazeer2020glu} activation functions, and RMS Normalization. It utilizes a SentencePiece-based tokenizer with a compressed vocabulary of 32,000 tokens. The model was pre-trained on a combination of SlimPajama (deduplicated web data) and StarCoder (source code).

\paragraph{\texttt{Qwen}:} 
This one utilizes a modern Transformer architecture that also features RoPE, SwiGLU, and RMSNorm, but is distinguished by its byte-fallback BPE tokenizer with a vocabulary of 151,643 tokens. Unlike the other English-centric models, \texttt{Qwen} was pre-trained on a diverse, massive-scale multilingual web corpus.

\section{Implementation Details for Reproducibility}
\label{app:implementation}

\subsection{Execution Platform and Hardware}
All experiments were run in Kaggle notebooks using a runtime provisioned with two NVIDIA Tesla T4 GPUs (16~GiB VRAM per GPU). The main scoring notebooks checked CUDA availability at runtime and selected the GPU when available; GPU inference used FP16 model weights, with FP32 specified as the CPU fallback. The primary four-model pipeline loaded and evaluated one checkpoint at a time, then released the model and cleared the CUDA cache before loading the next checkpoint. It used inference-only execution and did not fine-tune model weights. Although the allocated runtime had two T4 GPUs, the primary four-model scoring code moved each model to the default CUDA device rather than explicitly sharding these models across both GPUs. Internet access was enabled in Kaggle for retrieving public Hugging Face checkpoints and the compression reference text when they were not already cached.

\subsection{PDF Processing and Data Handling}
PyMuPDF (imported as \texttt{fitz}) was used to extract text, page counts, layout blocks, and document metadata from the poetry and academic PDF collections. The project-specific Python code handled reading-order reconstruction, paragraph and stanza identification, heading and caption filters, and source-level chunk manifests. The academic-paper pipeline used SHA-256 hashes to identify exact duplicate PDFs and recorded input validation results before scoring. NumPy and pandas were used to maintain the chunk-level records, source-level summaries, and tabular manifests.

The main cross-model runs used 28 poets and 29 academic papers. Chunk text was passed unchanged in case and with tokenizer-added special tokens disabled. The saved run records include source and chunk identifiers, token and character counts, scoring status, context length, and the model revision when provided by the Hugging Face configuration. The paper-analysis notebook expected exactly 29 unique readable PDFs and stopped if the validated count differed.

\subsection{Language-Model Inference}
The language-model interface was provided by Hugging Face Transformers through \texttt{AutoTokenizer} and \texttt{AutoModelForCausalLM}; PyTorch performed the forward passes and loss calculations. The following public checkpoint identifiers were used in the primary four-model comparison.

\begin{table}[h]
\centering
\small
\resizebox{\columnwidth}{!}{%
\begin{tabular}{p{0.3\columnwidth}p{0.68\columnwidth}}
\hline
\textbf{Model} & \textbf{Checkpoint identifier} \\
\hline
GPT-2 & \texttt{gpt2} \\
Pythia-410M & \texttt{EleutherAI/pythia-410m} \\
Qwen2.5-0.5B & \texttt{Qwen/Qwen2.5-0.5B} \\
TinyLlama-1.1B & \texttt{TinyLlama/TinyLlama-1.1B-Chat-v1.0} \\
\hline
\end{tabular}%
}
\caption{Language-model checkpoints used in the experiments. All models were used for language model scoring purpose.}
\label{tab:implementation-models}
\end{table}

The scoring code used PyTorch inference mode and a stride of 512 tokens, with the usable context length inferred from the tokenizer and model configuration. Models were loaded with remote-code execution disabled. The scorer emitted bits per character (BPC), bits per token, token counts, and character counts for each chunk; the source-level output also retained the scoring status and contextual metadata needed for auditing. No autoregressive text generation was used for the primary scores.

\subsection{Compression-Based Baselines}
The supplemental compression-baseline notebooks used the Python standard-library implementations of \texttt{DEFLATE}, \texttt{LZMA}, and \texttt{bzip2}, together with the third-party \texttt{zstandard} package for Zstandard compression. The compressor settings used in the code are summarized in Table~\ref{tab:implementation-compressors}.

\begin{table}[h]
\centering
\small
\begin{tabular}{p{0.30\linewidth}p{0.58\linewidth}}
\hline
\textbf{Baseline} & \textbf{Implementation setting} \\
\hline
Plain DEFLATE & \texttt{zlib}; raw DEFLATE stream, level 9 \\
Plain LZMA & \texttt{lzma}; raw LZMA2 stream, preset 9 with the extreme preset \\
Plain bzip2 & \texttt{bz2}; compression level 9 \\
Plain Zstandard & \texttt{zstandard}; compressor level 19 \\
Primed DEFLATE & Difference between compressed reference-plus-chunk size and compressed reference size, using raw DEFLATE level 9 \\
Primed LZMA & The analogous size difference using raw LZMA2, preset 9 with the extreme preset \\
\hline
\end{tabular}
\caption{Compression implementations and settings used as non-neural baselines.}
\label{tab:implementation-compressors}
\end{table}

The primed baselines used an 80,000-character prefix taken from the body of \textit{The Adventures of Sherlock Holmes}, Project Gutenberg eBook~1661. The text was retrieved with \texttt{requests}, saved to the notebook's working directory, and encoded as UTF-8 before compression. Compressed byte lengths were converted to bits and normalized by the number of text characters, rather than by the number of UTF-8 bytes. The compression runs reused the four language-model checkpoint identifiers listed in Table~\ref{tab:implementation-models} to provide a common set of source chunks for the two types of baseline.

\subsection{Numerical Analysis, Correlations, and Outputs}
NumPy was used for histogram construction and numerical moment calculations; the Freedman--Diaconis rule was requested through \texttt{numpy.histogram\_bin\_edges}. For the binned source-level summaries, the proposed bin count was capped to the range 8--30, and the saved outputs retained the nonempty bin centres and their frequencies. Pandas was used to reshape and export the resulting tables. SciPy's statistical functions supplied the pairwise Spearman rank correlations, Kendall and Pearson correlations. \texttt{Matplotlib} was used for figures.

The compression-baseline notebook calculated both raw-chunk and binned-distribution moments. Its comparison of average language-model--language-model correlations against compressor--language-model correlations used 5,000 source-level bootstrap resamples, configured with random seed 20261012, to form percentile intervals for the difference between the averages. The notebook also exported the pairwise correlations and diagnostic tables for the source-level associations involving mean BPC, variance, and chunk length.

\subsection{Library and Runtime Versions}
Table~\ref{tab:implementation-versions} records the software versions available from the saved notebook metadata and printed run output. Where the notebooks did not capture an exact installed version, it is identified as unrecorded rather than inferred from a different environment.

\begin{table}[h]
\centering
\footnotesize
\begin{tabular}{ll}
\hline
\textbf{Component} & \textbf{Version} \\
\hline
Python (analysis / compression) & 3.13.15 / 3.10.0 \\
PyTorch & 2.11.0+cu128 \\
PyMuPDF & 1.28.2 \\
Transformers & $\geq 4.45$, $<5$ (constraint) \\
NumPy, pandas, SciPy & Not recorded \\
Matplotlib, Accelerate & Not recorded \\
Zstandard, requests & Not recorded \\
\hline
\end{tabular}
\caption{Software versions recorded in the notebooks.}
\label{tab:implementation-versions}
\end{table}

The saved notebooks do not pin every dependency or record a complete \texttt{pip freeze} for each Kaggle session. Consequently, the exact versions of the unrecorded packages, as well as the NVIDIA driver version, cannot be established reliably from the available run artifacts.
\end{document}